\pdfoutput=1

\documentclass[11pt]{article}

\usepackage[]{acl}

\usepackage{times}
\usepackage{latexsym}
\usepackage{multirow}
\usepackage{graphicx}
\usepackage{booktabs}
\usepackage{enumitem}
\usepackage{algorithm}
\usepackage{algorithmic}
\usepackage{amsmath}
\usepackage{amssymb}
\usepackage{bm}
\usepackage{colortbl}

\usepackage{times}
\usepackage{latexsym}
\usepackage{tikz}
\usepackage{graphicx}
\usepackage[edges]{forest}
\usepackage{custom}
\usepackage{scalefnt}
\usepackage[most]{tcolorbox}
\usepackage[framemethod=tikz]{mdframed}
\usepackage{subcaption}
\usepackage{soul}
\usepackage{pifont}
\usepackage{xcolor}

\definecolor{hidden-black}{RGB}{20,68,106}

\usepackage[T1]{fontenc}

\usepackage[utf8]{inputenc}

\usepackage{microtype}

\usepackage{inconsolata}

\title{RSIPO: Recursive Self-Improving Policy Optimization for Agentic Reinforcement Learning via Skill-Policy Co-Evolution}

\makeatletter
\renewcommand{\@fnsymbol}[1]{}
\makeatother

\author{Fan Zhang{$^{1*}$},~Hao Wu{$^{2*}$},~Jinyang Wu{$^2$},~Hongyu Qu{$^3$},~Zheng Lian$^4$,\\ \textbf{~Xiao Luo{$^5$},~Yefeng Zheng$^6$,~Bo Zhang$^{7\dagger}$,~Haoxuan Li$^{8\dagger}$\thanks{$^*$Equal Contribution. $^\dagger$Corresponding Authors.},~Pheng-Ann Heng$^1$} \\
{$^1$CUHK} \quad {$^2$THU} \quad {$^3$NUS} \quad {$^4$Tongji University} \quad {$^5$UW-Madison}\\ \quad {$^6$Westlake University} \quad {$^7$Shanghai AI Laboratory} \quad {$^8$PKU}
}

\begin{document}
\maketitle

\begin{abstract}
Recent focus in agentic reinforcement learning (RL) seeks to improve credit assignment in long-horizon tasks. One promising direction combines reinforcement learning with verifiable rewards (RLVR) and on-policy self-distillation (OPSD), integrating trajectory-level advantages with dense token-level supervision. This line of work often conditions a teacher model on privileged context, such as agentic skills, to score tokens generated by the student policy. Despite their progress, they largely overlook \ding{182} the on-policy evolution of agentic skills and \ding{183} the reliability of teacher supervision. Towards this end, we introduce Recursive Self-Improving Policy Optimization (RSIPO), a self-evolving agentic RL framework that enables agentic skills and the policy model to co-evolve throughout the training process. Specifically, RSIPO enables the policy itself to extract skills from completed trajectories and explicitly optimizes its skill-generation capability through trajectory replay. It further employs empirical Bayes shrinkage to dynamically calibrate the student's trust in teacher scores, accounting for the fact that skill-enhanced teacher guidance can sometimes be less reliable. Experiments with Qwen-family models on embodied interaction and search-based agentic tasks demonstrate that RSIPO outperforms all the compared baseline approaches. With Qwen2.5-7B as the base model, RSIPO achieves an average success rate of 98.6\% on ALFWorld, yielding absolute improvements of 2.5\% and 17.4\% over the second-best baseline and standard GRPO, respectively.
Code is available at \url{https://github.com/zfkarl/RSIPO}.
\end{abstract}

\section{Introduction}

Agentic reinforcement learning (RL) \cite{zhang2025landscape,zhang2025agentrl} has emerged as a key component of large language model (LLM) post-training, substantially improving model performance on long-horizon multi-turn tasks. Unlike conventional RL for static and single-turn scenarios \cite{guo2025deepseek}, agentic RL aims to train models through interactions with dynamic environments, where they iteratively reason, invoke tools, and incorporate observations across multi-turn conversations \cite{lu2026skill0,singh2025agentic}. This requires learning sequential decision-making policies that reliably guide agents toward successful task completion.

\begin{figure*}[t]
\centering
\includegraphics[width=\textwidth]{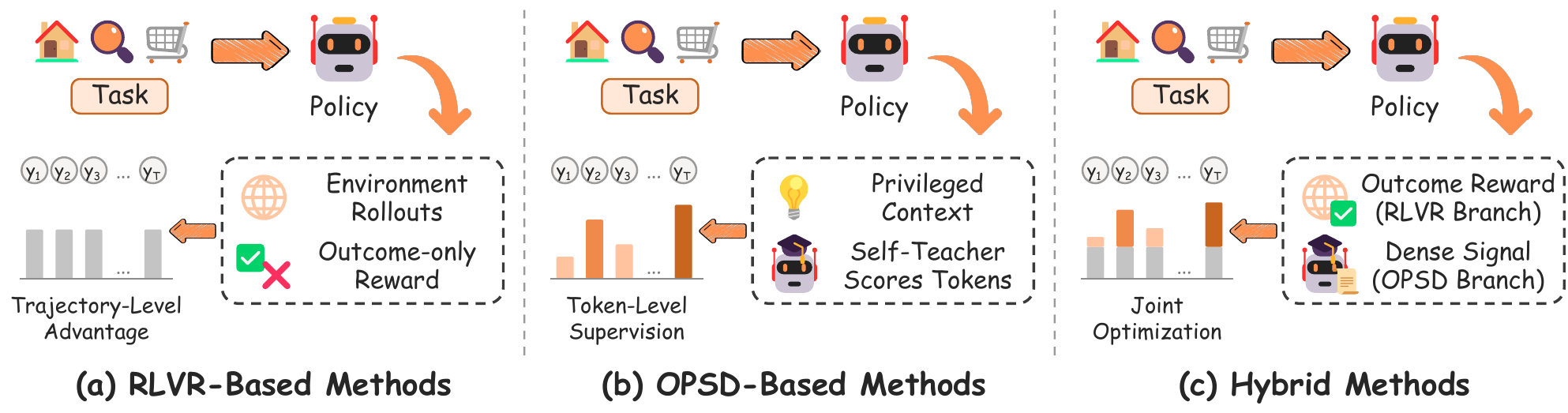}
\caption{Comparison of three paradigms of agentic RL. \textbf{(a)} RLVR-based methods assign equal trajectory-level advantages to all response tokens. \textbf{(b)} OPSD-based methods provide token-level supervision from a self-teacher. \textbf{(c)} Hybrid methods combine the two branches, typically using agentic skills as the teacher's privileged context.}
\label{fig:paradigms}
\end{figure*}

As shown in Figure~\ref{fig:paradigms}, recent agentic RL approaches have explored reinforcement learning with verifiable rewards (RLVR), on-policy self-distillation (OPSD), and hybrid training frameworks that combine the two. Among them, RLVR-based methods provide effective outcome-based trajectory-level advantages, whereas OPSD-based methods offer dense and fine-grained token-level supervision. Hybrid methods seek to combine these complementary signals, typically coupling an RLVR branch with an OPSD branch for joint optimization. In the OPSD branch, a teacher model conditioned on privileged context scores student-generated tokens, and the student is trained to reduce the token-level log-probability gap between the two models. For example, SDAR \cite{lu2026self} uses agentic skills retrieved from an external skill library as privileged context for the teacher model, whereas OPID \cite{yang2026opid} and SEED \cite{wu2026seed} explore generating such skills with external or internal analyzers.

\begin{figure*}[t]
\centering
\includegraphics[width=\textwidth]{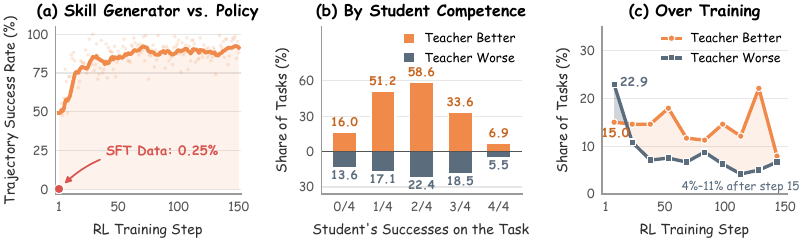}
\caption{Preliminary analysis with Qwen2.5-7B-Instruct on ALFWorld. \textbf{(a)} Success rate of the skill generator rollouts on SFT data and of the policy rollouts during RL training. \textbf{(b)} Share of training tasks on which the skill-enhanced teacher succeeds more or less often than the student, grouped by the student's successes in four separate rollouts on the task. \textbf{(c)} The same shares over RL training, averaged over blocks of 15 steps.}
\label{fig:intro_evidence}
\end{figure*}

Despite their progress in advancing agentic RL, existing hybrid approaches face two key limitations. First, agentic skills are either retrieved from a static external library or produced by a skill generator trained off-policy in advance. Consequently, while the policy improves at task execution during training, skill generation is not explicitly optimized on-policy to evolve alongside it. Following the experimental setup of SEED \cite{wu2026seed}, we find that only 0.25\% of the trajectories in the supervised fine-tuning (SFT) data used to train the skill generator are successful, mainly because the base model follows the required action format in just 11.0\% of its steps, whereas the policy's rollout success rate rapidly exceeds 80\% during subsequent RL training (Figure~\ref{fig:intro_evidence} (a)). As a result, 86.8\% of the skills in the SFT data concern action format or syntax, even though 99.7\% of the policy's actions are already valid in the first 10 RL steps. The trajectories that the skill generator must summarize thus shift from almost exclusively failures to predominantly successes, creating a distribution shift that its initial off-policy training does not explicitly address.
Second, the OPSD objective encourages the student to align with a skill-enhanced teacher, implicitly assuming that the teacher consistently provides more reliable guidance than the student's own decisions. However, access to skills does not guarantee better performance. As shown in Figure~\ref{fig:intro_evidence} (b), the skill-enhanced teacher underperforms the student on 5.5\%--22.4\% of tasks across different levels of student competence. Its advantage is most pronounced on tasks that the student solves in one to three of four rollouts. By contrast, for tasks with zero or four successful student rollouts, the proportions of tasks on which the teacher outperforms and underperforms the student differ only slightly (16.0\% vs.\ 13.6\% and 6.9\% vs.\ 5.5\%, respectively). Figure~\ref{fig:intro_evidence} (c) further shows that, during the first 15 RL training steps, the teacher underperforms the student on more tasks than it outperforms the student (22.9\% vs.\ 15.0\%) and continues to underperform on 4\%--11\% of tasks thereafter. Therefore, uniformly trusting teacher guidance exposes the student policy to unreliable supervision.

These observations motivate Recursive Self-Improving Policy Optimization (RSIPO), a self-evolving agentic RL framework that jointly improves skill generation and task execution while calibrating the student's trust in teacher supervision. Building on the RLVR and OPSD branches, RSIPO uses the student policy to rewrite skills and replays half of its rollouts with these skills, measuring the resulting gains relative to the remaining half, which serves as a control group. Each replay gain is centered by subtracting the mean gain for the same task type, and the resulting reward is assigned to the generated skill tokens. This explicitly optimizes the policy's skill-generation capability on-policy without requiring auxiliary judge models or additional evaluation data. Beyond skill evolution, RSIPO dynamically adjusts the student's reliance on the teacher in the OPSD branch. Specifically, it applies empirical Bayes shrinkage to the replay gains to obtain robust and task-dependent trust coefficients that weight the OPSD objective. These coefficients increase the student's reliance on reliable teacher guidance, downweight unreliable supervision, and can disable distillation entirely on tasks where the teacher is particularly unreliable.

We conduct extensive experiments across multiple base models on ALFWorld \cite{shridhar2020alfworld} for embodied interaction and seven datasets for search-based agentic tasks. The results show that RSIPO substantially outperforms a wide range of baselines, including RLVR-based, OPSD-based, and hybrid agentic RL methods. On ALFWorld, RSIPO achieves average task success rates of 95.7\% and 98.6\% with Qwen2.5-3B and Qwen2.5-7B \cite{qwen2025qwen25technicalreport}, respectively, surpassing the strongest prior method, SEED \cite{wu2026seed}, by 3.9\% and 2.5\%. The performance improvements over the standard GRPO+OPSD baseline are even larger, reaching 14.5\% and 18.2\%, respectively.

To summarize, our main contributions of this paper are threefold:
\begin{itemize}[leftmargin=*]
    \item[\ding{182}] \textbf{Co-Evolving Perspective.} We identify limitations in skill generation and teacher supervision in existing agentic RL methods and propose a co-evolving perspective that couples on-policy skill improvement with adaptive trust in the teacher.
    \item[\ding{183}] \textbf{Self-Evolving Framework.} We introduce RSIPO, a self-evolving framework that uses skill-induced performance gains measured through trajectory replay to drive skill-policy co-evolution and applies empirical Bayes shrinkage to calibrate the student's trust in teacher supervision.
    \item[\ding{184}] \textbf{Empirical Validation.} Extensive experiments across multiple models and datasets demonstrate that RSIPO outperforms existing agentic RL methods, achieving state-of-the-art performance on ALFWorld and search-based agentic tasks.
\end{itemize}

\section{Related Work}

\subsection{Agentic Reinforcement Learning}

Reinforcement learning with verifiable rewards (RLVR) has demonstrated substantial gains in single-turn reasoning. Critic-free methods such as GRPO~\cite{guo2025deepseek} and DAPO~\cite{yu2025dapo} estimate advantages from group-normalized outcome rewards, eliminating the need for the learned value network used in PPO~\cite{schulman2017proximal}. In agentic settings, policies interact with embodied~\cite{shridhar2020alfworld}, shopping~\cite{yao2022webshop}, or search~\cite{jin2025search} environments over multiple turns. Attributing a single terminal reward to a long sequence of decisions poses a credit assignment challenge, motivating multi-turn training frameworks~\cite{wang2025ragen,zhang2025agentrl} and step-level advantage estimation~\cite{feng2025group}. A complementary line of work equips agents with reusable natural-language experience, including reflections, insights, and skill libraries~\cite{shinn2023reflexion,zhao2024expel,wang2023voyager}, with recent methods incorporating such skills into agentic RL training~\cite{xia2026skillrl,lu2026skill0}. These approaches, however, typically rely on fixed skill libraries, frozen external generators, or skill-generation objectives that do not explicitly account for the utility of generated skills in task execution.

\subsection{On-Policy Self-Distillation}

\begin{figure*}[t]
\centering
\includegraphics[width=\textwidth]{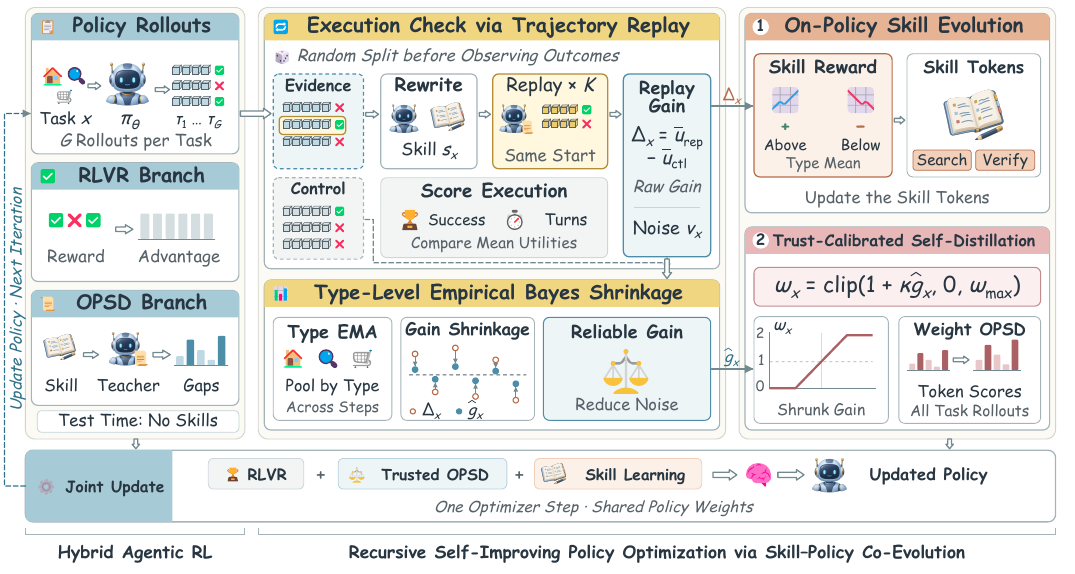}
\caption{Overview of the proposed RSIPO framework. Replay gains measured by the execution check reward on-policy skill generation and calibrate the student's trust in the skill-enhanced teacher.}
\label{fig:method}
\end{figure*}

In on-policy distillation (OPD), a teacher model scores sequences sampled from the student policy, allowing the student to learn from its own generation distribution rather than a fixed corpus~\cite{agarwal2024onpolicy,gu2024minillm}. On-policy self-distillation (OPSD) uses the same model as the teacher by conditioning it on privileged training-time context unavailable to the student, such as worked solutions or feedback on the student's attempts~\cite{zhao2026self,hubotter2026reinforcement}. Hybrid approaches combine this supervision with RLVR, often using agentic skills as privileged context for the teacher~\cite{wang2026skill}. For example, SDAR retrieves skills from an external library~\cite{lu2026self}, OPID uses an external analyzer to extract skills from completed episodes~\cite{yang2026opid}, and SEED trains the policy itself to generate skills through supervised fine-tuning~\cite{wu2026seed}. Other methods modify how the teacher--student log-probability gap contributes to optimization, for example by aggregating it at the action-step level or annealing the distillation signal over training~\cite{zhang2026stepopsd,wang2026agentopsd,tan2026atod}.

\section{Methodology}

\subsection{Preliminary and Notation}

We consider an agent that solves a task $x$ by interacting with an environment for at most $T_{\max}$ turns. At turn $t$, the policy $\pi_\theta$ receives a context $c_t$ that contains the task instruction, a window of recent observations and actions, and the current observation. Then it generates a response $y_t=(y_{t,1},\dots,y_{t,|y_t|})$ consisting of reasoning and action tokens. The environment executes the action and returns the next observation. An episode produces a trajectory $\tau=(c_1,y_1,\dots,c_T,y_T)$ with $T\le T_{\max}$ and a verifiable outcome reward $R(\tau)\in\{0,1\}$ that indicates task execution success.

\paragraph{RLVR Branch.} For each task, $G$ trajectories $\{\tau^{(i)}\}_{i=1}^{G}$ are sampled from the current policy. Following GRPO~\cite{shao2024deepseekmath}, trajectory $\tau^{(i)}$ receives the group-normalized advantage $A^{(i)}=\bigl(R(\tau^{(i)})-\mu_R\bigr)/\sigma_R$, where $\mu_R$ and $\sigma_R$ are the mean and standard deviation of the rewards in the group, and this advantage is shared by all tokens of $\tau^{(i)}$. The policy is optimized with the clipped surrogate objective $\mathcal{L}_{\text{PPO}}$~\cite{schulman2017proximal} and a KL penalty $\mathcal{L}_{\text{KL}}$ toward a reference policy:
\begin{equation}
\mathcal{L}_{\text{RLVR}}=\mathcal{L}_{\text{PPO}}+\lambda_{\text{KL}}\mathcal{L}_{\text{KL}}.\label{eq:rlvr}
\end{equation}

\paragraph{OPSD Branch.} Following SEED~\cite{wu2026seed}, the policy itself summarizes each completed trajectory into an agentic skill $s$, prompted with an analysis template that presents the trajectory and asks for an episode summary and a reusable skill. The skill-enhanced teacher is the same policy with the skill inserted into its context, $\pi_T(\cdot\mid c_t)=\pi_\theta(\cdot\mid c_t\oplus s)$. For each token generated by the student, the two log-probabilities define a gated distillation term:
\begin{align}
\ell_{t,j}&=g_{t,j}\bigl(\log\pi_T(y_{t,j})-\log\pi_\theta(y_{t,j})\bigr),\label{eq:opsd_term}\\
g_{t,j}&=\sigma\bigl(\beta\bigl(\log\pi_T(y_{t,j})-\operatorname{sg}[\log\pi_\theta(y_{t,j})]\bigr)\bigr),\nonumber
\end{align}
where the conditioning on $c_t$ and $y_{t,<j}$ is omitted, $\operatorname{sg}[\cdot]$ stops gradients, and the teacher log-probabilities are treated as constants. The gate $g_{t,j}$ focuses distillation on tokens for which the teacher is more confident than the student. Let $\mathcal{L}_{\text{OPSD}}$ denote the average of $\ell_{t,j}$ over all response tokens, the hybrid objective is:
\begin{equation}
\mathcal{L}_{\text{hybrid}}=\mathcal{L}_{\text{RLVR}}+\lambda_{\text{OPSD}}\mathcal{L}_{\text{OPSD}}.\label{eq:hybrid}
\end{equation}
Following SEED \cite{wu2026seed}, the policy is initialized by supervised fine-tuning on skills that an external LLM writes for the base model's own trajectories, so that it can generate skills before RL begins. No skill is provided at test time.

\subsection{The Proposed RSIPO}

RSIPO keeps both RLVR and OPSD branches of the hybrid agentic RL framework and adds an execution check that measures how much a self-generated skill actually improves the policy's execution for every training task. The measured gains are used twice: \ding{182} they reward the policy model for writing helpful skills, so that skill generation co-evolves with task execution, and \ding{183} they calibrate the student's trust in the skill-enhanced teacher. 
Figure~\ref{fig:method} illustrates the overall RSIPO framework.

\paragraph{Execution Check via Trajectory Replay.}
Before any outcome is inspected, the $G$ rollouts of task $x$ are split uniformly at random into an evidence set $\mathcal{E}_x$ and a control set $\mathcal{C}_x$ of equal size, so that $\mathcal{C}_x$ is an unbiased sample of the student's current performance. RSIPO picks one trajectory from $\mathcal{E}_x$, preferring a successful one when available, and lets the policy rewrite a skill $s_x$ from it with the analysis template. If $s_x$ is valid, the skill-enhanced teacher $\pi_\theta(\cdot\mid\cdot\oplus s_x)$ is replayed $K$ times from exactly the same initial state of the task. Each episode is scored by the utility:
\begin{equation}
u(\tau)=\mathbb{I}\bigl[R(\tau)=1\bigr]\Bigl(1-\alpha\,\frac{T(\tau)}{T_{\max}}\Bigr),\label{eq:utility}
\end{equation}
where $T(\tau)$ is the number of turns and $\alpha$ favors faster successes. The replay gain compares the $K$ replays $\{\tilde\tau_k\}_{k=1}^{K}$ with the control rollouts:
\begin{equation}
\Delta_x=\frac{1}{K}\sum_{k=1}^{K}u(\tilde\tau_k)-\frac{1}{|\mathcal{C}_x|}\sum_{\tau\in\mathcal{C}_x}u(\tau).\label{eq:gain}
\end{equation}
Since both the replay and control groups contain only a few episodes, $\Delta_x$ could be noisy. We approximate its sampling variance by $v_x=\sum_{a\in\{\text{rep},\,\text{ctl}\}}\tilde u_a(1-\tilde u_a)/n_a$, where $n_a$ is the number of episodes in group $a$ and $\tilde u_a=(n_a\bar u_a+1)/(n_a+2)$ is its Laplace-smoothed mean utility.

\paragraph{Type-Level Empirical Bayes Shrinkage.}
To obtain reliable estimates from such noisy measurements, RSIPO pools the gains of tasks of the same type $c$, such as a task category in ALFWorld or a source dataset in search-based tasks, across training steps. For each type present in the current batch, it updates the moving averages as:
\begin{equation}
\mathbf{m}_c\leftarrow(1-\eta)\,\mathbf{m}_c+\eta\,\bigl(\overline{\Delta},\ \overline{\Delta^{2}},\ \overline{v},\ 1\bigr),\label{eq:ema}
\end{equation}
where the averages are taken over the tasks of type $c$ in the batch and the last entry corrects the bias of the zero initialization. Writing $\mathbf{m}_c=(m_c^{(1)},m_c^{(2)},m_c^{(v)},m_c^{(0)})$, the type mean $\bar\Delta_c=m_c^{(1)}/m_c^{(0)}$ and the between-task variance $\hat\sigma^2_c=\max\bigl(0,\,m_c^{(2)}/m_c^{(0)}-\bar\Delta_c^{2}-m_c^{(v)}/m_c^{(0)}\bigr)$, i.e., the total variance minus the noise variance, specify a Gaussian prior over the true gains of type $c$. Following empirical Bayes estimation~\cite{efron1975data}, the posterior mean of the gain of task $x$ shrinks its own estimate toward the type mean according to its reliability:
\begin{equation}
\hat g_x=w_x\Delta_x+(1-w_x)\bar\Delta_{c},\quad w_x=\frac{\hat\sigma^2_{c}}{\hat\sigma^2_{c}+v_x},\label{eq:shrink}
\end{equation}
with $c=c(x)$. Gains that are noisy or belong to types without genuine variation across tasks are pulled toward the type mean, whereas reliable and distinctive gains are largely preserved. When a task has no valid replay, the corresponding replay gain falls back to the prior mean $\hat g_x=\bar\Delta_{c}$.

\paragraph{On-Policy Skill Evolution.}
RSIPO uses the execution check to train skill generation directly. Each task contributes one sample, whose input is the analysis prompt $q_x$ of the selected evidence trajectory and whose output $y^{(x)}$ is the rewritten analysis containing $s_x$. The reward is defined as the replay gain centered by the type mean:
\begin{equation}
r_x=\Delta_x-\bar\Delta_{c(x)},\label{eq:reward}
\end{equation}
where $r_x$ is positive only if $s_x$ yields a larger replay gain than the policy's skills achieve on average for tasks of the same type, and negative otherwise. In contrast to the trust coefficient below, $r_x$ is not shrunk, because a policy-gradient signal should be unbiased and shrinkage would suppress all rewards whenever the between-task variance cannot be separated from noise. An invalid skill receives the penalty $r_x=-\rho$, and a task whose replay does not reproduce the initial state receives $r_x=0$.\footnote{For search-based tasks, invalid skills are not penalized ($\rho=0$), and a small reward of $0.05$ is assigned to the format tokens of valid outputs to keep the JSON structure stable.} The skill-generation loss is a reward-weighted log-likelihood over the tokens of the skill:
\begin{equation}
\bar\ell_x=\frac{\sum_{j}m_{x,j}\log\pi_\theta\bigl(y^{(x)}_j\mid q_x,y^{(x)}_{<j}\bigr)}{\max\bigl(1,\sum_j m_{x,j}\bigr)},\label{eq:skill_ll}
\end{equation}
\begin{equation}
\mathcal{L}_{\text{skill}}=-\frac{1}{|X|}\sum_{x\in X}r_x\,\bar\ell_x,\label{eq:skill}
\end{equation}
where $X$ is the task batch and $m_{x,j}=1$ marks the tokens inside the skill string of a valid output, or all tokens of an invalid one. Because the skills are written by the current policy from its own trajectories and rewarded by their effect on its own execution, skill generation is optimized on-policy and evolves together with task execution.

\paragraph{Trust-Calibrated Self-Distillation.}
The shrunk gain $\hat g_x$ estimates how much the skill-enhanced teacher outperforms the student on task $x$. RSIPO maps it to a trust coefficient:
\begin{equation}
\omega_x=\operatorname{clip}\bigl(1+\kappa\,\hat g_x,\ 0,\ \omega_{\max}\bigr),\label{eq:trust}
\end{equation}
which weights the OPSD terms of all rollouts:
\begin{equation}
\mathcal{L}_{\text{OPSD}}(\boldsymbol{\omega})=\frac{1}{N}\sum_{i}\sum_{t,j}\omega_{x(i)}\,\ell^{(i)}_{t,j},\label{eq:opsd_trust}
\end{equation}
where $\ell^{(i)}_{t,j}$ is the term in Eq.~\eqref{eq:opsd_term} for the $j$-th token of turn $t$ in rollout $i$, $x(i)$ is the rollout task, and $N$ is the number of response tokens. Setting $\omega_x=1$ for all tasks recovers the original OPSD loss. With $\kappa=2$ and $\omega_{\max}=2$, distillation is disabled when $\hat g_x\le-0.5$ and doubled when $\hat g_x\ge0.5$. We use $\hat g_x$ rather than $\Delta_x$ because a noisy per-task estimate would directly distort the loss weight.

\paragraph{Overall Objective.}
RSIPO optimizes all components in a single update:
\begin{equation}
\begin{aligned}
\mathcal{L}={}\mathcal{L}_{\text{RLVR}}+\lambda_{\text{OPSD}}\mathcal{L}_{\text{OPSD}}(\boldsymbol{\omega})+\mathcal{L}_{\text{skill}}.
\end{aligned}\label{eq:total}
\end{equation}
The gradient of $\mathcal{L}_{\text{skill}}$ is accumulated with the last PPO mini-batch so that both share one optimizer step, and the type statistics are committed only after the update succeeds. Algorithm~\ref{alg:rsipo} in Appendix~\ref{app:algorithm} summarizes one training step of RSIPO.

\section{Experiment}

\subsection{Experimental Setup}

\paragraph{Benchmarks.}
We evaluate the proposed RSIPO framework and baseline approaches on embodied interaction tasks using ALFWorld~\cite{shridhar2020alfworld} and search-based agentic tasks. Following Search-R1~\cite{jin2025search}, we employ three single-hop datasets, NQ~\cite{kwiatkowski2019natural}, TriviaQA~\cite{joshi2017triviaqa}, and PopQA~\cite{mallen2023trust}, and four multi-hop datasets, HotpotQA~\cite{yang2018hotpotqa}, 2WikiMultiHopQA~\cite{ho2020constructing}, MuSiQue~\cite{trivedi2022musique}, and Bamboogle~\cite{press2023measuring} for search-based tasks. We report success rates on ALFWorld and exact-match accuracy on search-based tasks.

\paragraph{Baselines.}
We compare RSIPO with 13 baselines, including training-free methods like Vanilla and Skill-Prompt*, and methods from the three paradigms in Figure~\ref{fig:paradigms}, namely the RLVR-based GRPO~\cite{shao2024deepseekmath}, Skill-GRPO, and Skill-GRPO*, the OPSD-based OPSD~\cite{zhao2026self}, and the hybrid GRPO+OPSD, Skill-SD~\cite{wang2026skill}, RLSD~\cite{yang2026rlsd}, SDAR~\cite{lu2026self}, StepOPSD~\cite{zhang2026stepopsd}, OPID~\cite{yang2026opid}, and SEED~\cite{wu2026seed}. Skill-Prompt* adds skills from the SkillRL library~\cite{xia2026skillrl} to the prompt at test time, while Skill-GRPO uses them during training and Skill-GRPO* also keeps them at test time. All other methods, including RSIPO, act without skills at test time.

\paragraph{Implementation Details.}
We employ various Qwen-family models, including Qwen2.5-3B-Instruct, Qwen2.5-7B-Instruct~\cite{qwen2025qwen25technicalreport}, and Qwen3-1.7B~\cite{yang2025qwen3} as base models. Following SEED \cite{wu2026seed}, we first fine-tune each model to generate agentic skills using trajectory-skill pairs that are annotated by GLM-5.2~\cite{zai2026glm52} on training data. RSIPO then runs for 150 steps with $G=8$ rollouts per task and 16 tasks per step on ALFWorld or 128 on search-based tasks, using a learning rate of $10^{-6}$, and a KL coefficient of 0.01 on ALFWorld and 0.001 on search-based tasks. We keep the OPSD setting with $\lambda_{\text{OPSD}}=0.01$ and $\beta=5$. For RSIPO, the evidence and control sets each contain half of the rollouts, and we set $K=4$, $\alpha=0.5$, $\eta=0.1$, and $\rho=0.2$. Each run uses 8 $\times$ NVIDIA H20 GPUs.

\subsection{Main Results}

\definecolor{topcolor}{RGB}{252,236,196}
\definecolor{secondcolor}{RGB}{223,235,253}
\newcolumntype{C}{>{\centering\arraybackslash}p{0.058\textwidth}}

\begin{table*}[!t]
\centering
\setlength{\tabcolsep}{3pt}
\caption{\textbf{Results on ALFWorld and search-based agentic tasks.} We report success rates (\%) on ALFWorld and exact-match accuracy (\%) on the seven search-based datasets. Methods marked with * indicate they use skills at test time. Within each base model, the \sethlcolor{topcolor}\hl{\textbf{best}} and \sethlcolor{secondcolor}\hl{\mbox{\underline{second-best}}} results are highlighted.}
\label{tab:main_results}
\resizebox{\textwidth}{!}{%
\begin{tabular}{l*{15}{C}}
\toprule
\multirow{2}{*}[-0.6ex]{\textbf{Method}} & \multicolumn{7}{c}{\textbf{ALFWorld}} & \multicolumn{8}{c}{\textbf{Search-Based Tasks}} \\
\cmidrule(lr){2-8} \cmidrule(lr){9-16}
& \textbf{Pick} & \textbf{Look} & \textbf{Clean} & \textbf{Heat} & \textbf{Cool} & \textbf{Pick2} & \textbf{Avg} & \textbf{NQ} & \textbf{Triv} & \textbf{Pop} & \textbf{Hotp} & \textbf{2Wk} & \textbf{MuS} & \textbf{Bam} & \textbf{Avg} \\
\midrule
\rowcolor{gray!10} \multicolumn{16}{c}{\textit{Qwen2.5-7B-Instruct}} \\
Vanilla
  & 36.1 & 22.2 & 3.1 & 0.0 & 0.0 & 0.0 & 12.5
  & 25.2 & 50.8 & 29.5 & 29.0 & 29.0 & 10.4 & 63.7 & 33.9 \\
Skill-Prompt*
  & 51.7 & 50.0 & 32.3 & 5.3 & 4.3 & 0.0 & 23.4
  & 30.9 & 52.1 & 32.7 & 32.7 & 27.9 & 12.7 & 66.1 & 36.4 \\
OPSD
  & 50.0 & 60.0 & 22.7 & 21.4 & 17.6 & 9.5 & 32.8
  & 8.8 & 8.6 & 17.5 & 2.5 & 4.2 & 0.5 & 1.2 & 6.2 \\
GRPO
  & 91.2 & 87.5 & 96.2 & 81.0 & 65.0 & 57.9 & 81.2
  & 45.1 & 63.7 & 44.0 & 43.6 & 43.2 & 16.8 & 37.6 & 42.0 \\
Skill-GRPO
  & 88.5 & 66.7 & 65.2 & 61.1 & 57.7 & 73.1 & 69.5
  & 45.2 & 63.7 & 45.7 & 43.1 & 43.3 & 19.6 & 21.4 & 40.3 \\
Skill-GRPO*
  & \cellcolor{topcolor}\textbf{100.0} & 83.3 & 96.4 & 83.3 & 75.0 & 78.9 & 88.3
  & 44.8 & 63.0 & 45.1 & 43.7 & 43.7 & 20.5 & 71.4 & 47.5 \\
GRPO+OPSD
  & 91.4 & 61.5 & \cellcolor{topcolor}\textbf{100.0} & 87.5 & 76.5 & 52.2 & 80.4
  & \cellcolor{secondcolor}\underline{47.3} & 64.5 & 46.9 & 43.8 & 39.3 & 18.0 & 69.4 & 47.0 \\
Skill-SD
  & 93.9 & \cellcolor{secondcolor}\underline{93.8} & 90.9 & \cellcolor{topcolor}\textbf{100.0} & 69.2 & 68.4 & 85.1
  & 47.1 & 64.5 & 47.8 & 44.2 & 42.1 & 20.2 & 69.0 & 47.8 \\
RLSD
  & \cellcolor{topcolor}\textbf{100.0} & 87.5 & 92.3 & 58.8 & 80.0 & 65.2 & 82.0
  & 46.8 & 63.0 & 44.4 & \cellcolor{secondcolor}\underline{45.5} & \cellcolor{secondcolor}\underline{48.9} & \cellcolor{secondcolor}\underline{21.5} & \cellcolor{secondcolor}\underline{73.0} & 49.0 \\
SDAR
  & 94.7 & 75.0 & \cellcolor{topcolor}\textbf{100.0} & 86.7 & 68.2 & 78.9 & 85.9
  & 46.3 & 63.5 & \cellcolor{secondcolor}\underline{48.2} & 43.8 & 48.4 & 19.6 & \cellcolor{secondcolor}\underline{73.0} & 49.0 \\
StepOPSD
  & \cellcolor{secondcolor}\underline{98.1} & 75.0 & \cellcolor{topcolor}\textbf{100.0} & \cellcolor{secondcolor}\underline{90.5} & 80.0 & 63.2 & 88.4
  & 45.3 & 64.6 & 45.1 & 44.5 & 44.4 & 19.3 & 69.8 & 48.2 \\
OPID
  & \cellcolor{topcolor}\textbf{100.0} & 81.8 & \cellcolor{secondcolor}\underline{97.1} & \cellcolor{topcolor}\textbf{100.0} & 80.8 & \cellcolor{secondcolor}\underline{80.0} & 90.0
  & \cellcolor{topcolor}\textbf{48.8} & \cellcolor{secondcolor}\underline{65.6} & 46.8 & \cellcolor{topcolor}\textbf{46.1} & 42.7 & \cellcolor{topcolor}\textbf{21.7} & 72.6 & \cellcolor{secondcolor}\underline{49.2} \\
SEED
  & \cellcolor{topcolor}\textbf{100.0} & \cellcolor{topcolor}\textbf{100.0} & 96.3 & 80.0 & \cellcolor{topcolor}\textbf{100.0} & \cellcolor{topcolor}\textbf{100.0} & \cellcolor{secondcolor}\underline{96.1}
  & 47.0 & 64.9 & 47.8 & 45.2 & 45.3 & 20.1 & 70.2 & 48.6 \\
\textbf{RSIPO}
  & \cellcolor{topcolor}\textbf{100.0} & \cellcolor{topcolor}\textbf{100.0} & 96.3 & \cellcolor{topcolor}\textbf{100.0} & \cellcolor{secondcolor}\underline{96.0} & \cellcolor{topcolor}\textbf{100.0} & \cellcolor{topcolor}\textbf{98.6}
  & 46.8 & \cellcolor{topcolor}\textbf{67.8} & \cellcolor{topcolor}\textbf{49.1} & 45.3 & \cellcolor{topcolor}\textbf{52.1} & 20.8 & \cellcolor{topcolor}\textbf{73.8} & \cellcolor{topcolor}\textbf{50.8} \\
\midrule
\rowcolor{gray!10} \multicolumn{16}{c}{\textit{Qwen3-1.7B}} \\
Vanilla
  & 25.0 & 22.2 & 3.1 & 0.0 & 21.4 & 4.2 & 12.5
  & 29.4 & 46.9 & 37.0 & 23.5 & 19.6 & 6.4 & 10.5 & 24.8 \\
Skill-Prompt*
  & 10.3 & 50.0 & 16.1 & 0.0 & 0.0 & 5.0 & 9.4
  & 29.4 & 46.5 & 36.2 & 22.9 & 20.8 & 4.3 & 10.1 & 24.3 \\
OPSD
  & 26.3 & 33.3 & 9.1 & 0.0 & 4.5 & 5.3 & 14.1
  & 4.2 & 8.3 & 4.6 & 6.6 & 15.3 & 0.7 & 1.2 & 5.8 \\
GRPO
  & 71.1 & 41.7 & 36.4 & 40.0 & 31.8 & 31.6 & 46.1
  & 40.0 & \cellcolor{topcolor}\textbf{58.9} & 43.5 & 35.4 & 30.3 & 12.0 & 65.7 & 40.8 \\
Skill-GRPO
  & 27.6 & 54.5 & 22.7 & 27.3 & 0.0 & 19.2 & 21.1
  & 39.2 & \cellcolor{secondcolor}\underline{58.6} & 43.9 & 35.2 & 28.2 & 11.5 & \cellcolor{secondcolor}\underline{66.1} & 40.4 \\
Skill-GRPO*
  & 31.4 & 42.9 & 51.9 & 8.3 & 11.5 & 7.1 & 28.1
  & 38.0 & 58.4 & 43.9 & 36.3 & 29.0 & 12.5 & \cellcolor{topcolor}\textbf{66.9} & 40.7 \\
GRPO+OPSD
  & 38.2 & 50.0 & 30.8 & 28.6 & 30.0 & 21.1 & 32.0
  & \cellcolor{secondcolor}\underline{40.7} & \cellcolor{topcolor}\textbf{58.9} & 45.0 & \cellcolor{topcolor}\textbf{37.0} & 34.6 & \cellcolor{topcolor}\textbf{13.3} & 65.7 & \cellcolor{secondcolor}\underline{42.2} \\
Skill-SD
  & 52.9 & 37.5 & 69.2 & 42.9 & 60.0 & 36.8 & 52.3
  & 39.1 & 57.5 & 45.4 & 34.8 & 34.1 & 10.7 & 64.1 & 40.8 \\
RLSD
  & 50.0 & 37.5 & 61.5 & 19.0 & 50.0 & 21.1 & 42.2
  & 38.6 & 57.3 & 43.0 & 34.5 & 34.1 & 11.5 & 65.3 & 40.6 \\
SDAR
  & 73.5 & 25.0 & 76.9 & 33.3 & 40.0 & 36.8 & 53.9
  & 39.7 & \cellcolor{topcolor}\textbf{58.9} & 45.3 & 35.9 & \cellcolor{secondcolor}\underline{35.5} & \cellcolor{secondcolor}\underline{12.6} & 65.3 & 41.9 \\
StepOPSD
  & 64.7 & 44.4 & 56.5 & 60.9 & 42.1 & 55.0 & 56.3
  & 40.5 & \cellcolor{secondcolor}\underline{58.6} & 45.6 & 34.9 & 29.8 & 10.8 & 64.5 & 41.4 \\
OPID
  & 65.9 & \cellcolor{secondcolor}\underline{72.7} & 66.7 & 40.0 & 63.2 & 45.0 & 58.9
  & 38.1 & 58.1 & 43.4 & 35.5 & 31.7 & 11.7 & 64.5 & 40.4 \\
SEED
  & \cellcolor{secondcolor}\underline{97.6} & \cellcolor{topcolor}\textbf{100.0} & \cellcolor{topcolor}\textbf{100.0} & \cellcolor{secondcolor}\underline{80.0} & \cellcolor{secondcolor}\underline{84.2} & \cellcolor{topcolor}\textbf{90.0} & \cellcolor{secondcolor}\underline{92.0}
  & \cellcolor{topcolor}\textbf{42.0} & \cellcolor{topcolor}\textbf{58.9} & \cellcolor{secondcolor}\underline{47.1} & \cellcolor{secondcolor}\underline{36.9} & 35.2 & 10.6 & 64.9 & \cellcolor{secondcolor}\underline{42.2} \\
\textbf{RSIPO}
  & \cellcolor{topcolor}\textbf{100.0} & \cellcolor{topcolor}\textbf{100.0} & \cellcolor{secondcolor}\underline{96.3} & \cellcolor{topcolor}\textbf{100.0} & \cellcolor{topcolor}\textbf{88.0} & \cellcolor{secondcolor}\underline{87.5} & \cellcolor{topcolor}\textbf{95.0}
  & 40.0 & 58.5 & \cellcolor{topcolor}\textbf{49.0} & 35.2 & \cellcolor{topcolor}\textbf{37.8} & \cellcolor{secondcolor}\underline{12.6} & 64.9 & \cellcolor{topcolor}\textbf{42.6} \\
\bottomrule
\end{tabular}}
\end{table*}

Table~\ref{tab:main_results} and Table~\ref{tab:main_results_qwen25_3b} compare RSIPO with all baselines on three base models. From these results, we can derive the following observations.

\paragraph{Embodied Interaction.}
On ALFWorld, RSIPO achieves the highest average success rate across all three backbones, reaching 95.7\%, 98.6\%, and 95.0\% with Qwen2.5-3B, Qwen2.5-7B, and Qwen3-1.7B, respectively. These results correspond to gains of 3.9\%, 2.5\%, and 3.0\% over the strongest baseline, SEED, and 20.7\%, 17.4\%, and 48.9\% over standard GRPO. Adding OPSD to GRPO does not consistently improve performance, as GRPO+OPSD underperforms GRPO with both Qwen2.5-7B and Qwen3-1.7B. This finding is consistent with our preliminary analysis in Figure~\ref{fig:intro_evidence}, which suggests that uniformly trusting teacher guidance can hinder student learning. In contrast, RSIPO outperforms GRPO+OPSD by 14.5\%, 18.2\%, and 63.0\% with Qwen2.5-3B, Qwen2.5-7B, and Qwen3-1.7B, respectively.

\paragraph{Search-Based Agentic Tasks.}
RSIPO attains the highest average accuracy with Qwen2.5-7B and Qwen3-1.7B, at 50.8\% and 42.6\%, and ranks second with Qwen2.5-3B, only 0.4\% below SEED. It improves over GRPO by 8.9\%, 8.8\%, and 1.8\% on the three backbones. The margins are smaller than on ALFWorld, likely because answering these questions relies more on retrieved evidence than on procedural skills, so the skill-enhanced teacher provides less additional guidance, as reflected by trust values close to one in Figure 4 (a).

\paragraph{Learning without Test-Time Skills.}
Despite using no agentic skills at test time for inference, RSIPO consistently outperforms baselines that receive skills during evaluation, including Skill-Prompt* and Skill-GRPO*, across both benchmarks and all evaluated backbones. With Qwen2.5-7B, RSIPO surpasses Skill-GRPO* by 10.3\% on ALFWorld and 3.3\% on search-based agentic tasks. These results suggest that trust-calibrated self-distillation helps the policy internalize knowledge from agentic skills generated during training, rather than relying on explicit skill guidance at inference.

\subsection{Further Analysis}

\begin{figure*}[t]
\centering
\includegraphics[width=\textwidth]{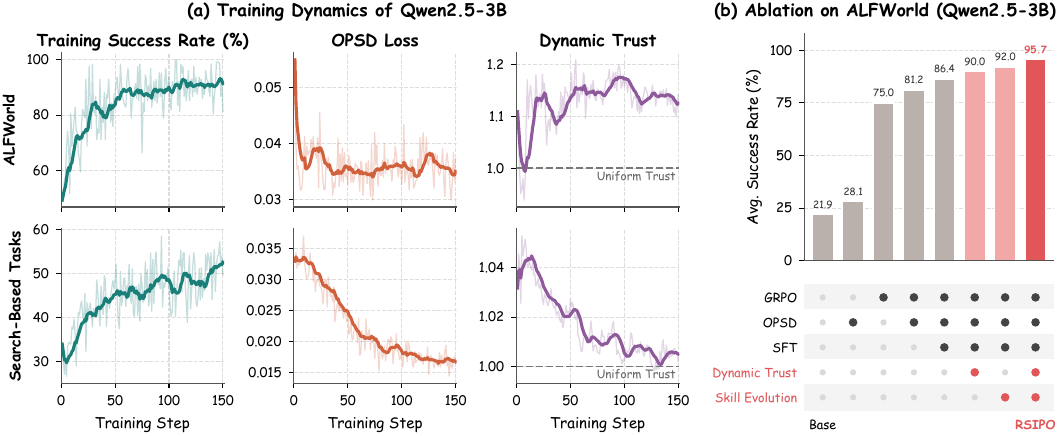}
\caption{\textbf{(a)} Training dynamics of RSIPO with Qwen2.5-3B-Instruct, where thick lines are 10-step moving averages. \textbf{(b)} Ablation on ALFWorld with Qwen2.5-3B-Instruct, where dots mark the components of each variant.}
\label{fig:dynamics_ablation}
\end{figure*}

\begin{figure*}[t]
\centering
\includegraphics[width=\textwidth]{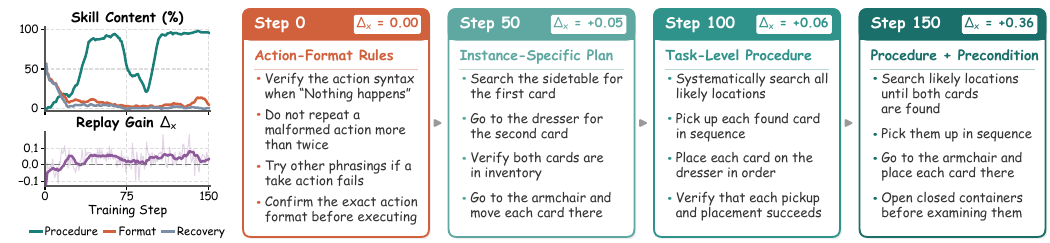}
\caption{On-policy skill evolution of Qwen2.5-7B-Instruct on ALFWorld. Left: skill content and replay gain $\Delta_x$. Right: condensed skills for tasks that place two credit cards, where step 0 is the SFT policy.}
\label{fig:skill_evolution}
\end{figure*}

\paragraph{Training Dynamics.}
Figure~\ref{fig:dynamics_ablation} (a) shows the training curves of RSIPO with Qwen2.5-3B. The training success rate rises from 63\% to over 90\% on ALFWorld and from 32\% to 53\% on search-based tasks, while the OPSD loss is roughly halved on search-based tasks and stays stable on ALFWorld. The dynamic trust tracks the reliability of the teacher. On ALFWorld, it exceeds 1.1 within 20 steps as the skill-enhanced teacher increasingly outperforms the control rollouts. On search-based tasks, where the teacher offers little extra benefit, it stays close to one, showing that even in such unfavorable cases the calibrated distillation degrades gracefully to approximately standard OPSD.

\paragraph{Ablation Study.}
Figure~\ref{fig:dynamics_ablation} (b) reports a cumulative ablation on ALFWorld with Qwen2.5-3B. OPSD alone improves the base model only from 21.9\% to 28.1\%, whereas GRPO reaches 75.0\%, and combining the two yields 81.2\%. With SFT initialization, the hybrid pipeline reaches 86.4\%. On top of it, dynamic trust and skill evolution bring gains of 3.6\% and 5.6\%, respectively, and the full RSIPO reaches 95.7\%. Removing either component from RSIPO costs 3.7\% to 5.7\%, showing that the two components are complementary.

\paragraph{Visualization of On-Policy Skill Evolution.}
Figure~\ref{fig:skill_evolution} traces the self-generated skills on ALFWorld using three non-exclusive keyword-based categories: Procedure (numbered steps), Format (action syntax), and Recovery (handling ``Nothing happens''). Early skills inherited from SFT often fall into Format or Recovery, with a negative average replay gain. Procedure skills then exceed 94\% in the last 45 steps, and the replay gain turns positive. For example, the step-0 skill only advises checking the syntax, whereas the step-150 skill specifies a systematic search-and-place procedure.

\section{Conclusion}

In this work, we presented RSIPO, a self-evolving agentic RL framework that enables agentic skills and the policy to co-evolve throughout training. RSIPO evaluates the utility of self-generated skills by comparing skill-enhanced teacher replays with control rollouts. The resulting replay gains reward the policy for generating useful skills and calibrate the student's trust in teacher supervision. Experiments on embodied and search-based tasks demonstrate RSIPO's advantages over existing agentic RL methods. Future work will extend RSIPO to larger models and a broader range of agentic tasks, including coding and co-working scenarios.

\section*{Limitations}

Despite its effectiveness, RSIPO has two main limitations. First, although trajectory replay requires neither additional evaluation data nor auxiliary evaluation models, the extra rollouts increase training costs. This introduces a trade-off between training effectiveness and computational efficiency. Second, RSIPO relies on environments that can be reset to a specified state, such as simulators and search environments backed by offline retrieval corpora. Extending RSIPO to non-resettable settings, such as interactions with live websites, remains a challenge, which we leave as future work.

\bibliography{custom}
\clearpage

\appendix
\label{appendix}

\section{Introduction to Baselines}
\label{app:baselines}

We compare the proposed RSIPO with a variety of baseline approaches, grouped by how they use outcome rewards and teacher supervision. Methods marked with * have access to skills at test time, whereas all other methods, including RSIPO, are evaluated without any skill in their context.

\paragraph{Training-free Methods.}
\textbf{Vanilla} evaluates the instruction-tuned base model without any post-training. \textbf{Skill-Prompt*} keeps the parameters of Vanilla fixed and inserts task-relevant skills retrieved from the SkillRL library~\cite{xia2026skillrl} into the prompt at test time, which measures the benefit of skills as pure in-context guidance.

\paragraph{RLVR-based Methods.}
\textbf{GRPO}~\cite{shao2024deepseekmath} normalizes the outcome rewards within a group of rollouts and assigns the resulting advantage to every token of a trajectory, without token-level supervision. \textbf{Skill-GRPO} optimizes the same objective but places retrieved agentic skills in the policy prompt during rollouts and updates, and removes them at test time to examine whether the skill-guided behavior is retained. \textbf{Skill-GRPO*} shares the training procedure of Skill-GRPO but keeps the skills at test time for evaluation.

\paragraph{OPSD-based Methods.}
\textbf{OPSD}~\cite{zhao2026self} uses the same model as student and teacher. The teacher additionally conditions on training-only privileged context and provides token-level targets on the student's own samples, while no privileged context is used at test time.

\begin{table}[t]
\centering
\footnotesize
\setlength{\tabcolsep}{4pt}
\caption{Hyperparameters on ALFWorld and search-based tasks. \textsuperscript{\dag}4096 for Qwen2.5-7B.}
\label{tab:hyperparameters}
\begin{tabular}{lcc}
\toprule
\textbf{Hyperparameter} & \textbf{ALFWorld} & \textbf{Search} \\
\midrule
\multicolumn{3}{l}{\textit{RL Training}} \\
Training Steps & 150 & 150 \\
Tasks per Step & 16 & 128 \\
Rollouts per Task $G$ & 8 & 8 \\
Learning Rate & $10^{-6}$ & $10^{-6}$ \\
Warmup Ratio & 0 & 0.1 \\
PPO Mini-Batch Size & 256 & 512 \\
PPO Clip Ratio & 0.2 & 0.2 \\
KL Coefficient $\lambda_{\text{KL}}$ & 0.01 & 0.001 \\
Entropy Coefficient & 0.001 & 0 \\
Invalid-Action Penalty & 0.1 & 0.01 \\
Max Prompt Length & 2048 & 4096 \\
Max Response Length per Turn & 512 & 512 \\
Max Turns $T_{\max}$ & 50 & 4 \\
History Length & 5 & 4 \\
Retrieved Passages per Query & -- & 3 \\
Training Temperature & 1.0 & 1.0 \\
Evaluation Temperature & 0.4 & 0.4 \\
\midrule
\multicolumn{3}{l}{\textit{OPSD Branch}} \\
Coefficient $\lambda_{\text{OPSD}}$ & 0.01 & 0.01 \\
Gate Sharpness $\beta$ & 5 & 5 \\
\midrule
\multicolumn{3}{l}{\textit{RSIPO}} \\
Evidence / Control Rollouts & 4 / 4 & 4 / 4 \\
Replays per Task $K$ & 4 & 4 \\
Skill Rewrite Temperature & 0.3 & 0.3 \\
Max Analysis Tokens & 4096 & 2048\textsuperscript{\dag} \\
Efficiency Weight $\alpha$ & 0.5 & 0.5 \\
EMA Rate $\eta$ & 0.1 & 0.1 \\
Trust Slope $\kappa$ & 2 & 2 \\
Max Trust $\omega_{\max}$ & 2 & 2 \\
Invalid-Skill Penalty $\rho$ & 0.2 & 0 \\
Format Reward for Valid Skills & -- & 0.05 \\
\bottomrule
\end{tabular}
\end{table}

\paragraph{Hybrid Methods.}
\textbf{GRPO+OPSD} adds the OPSD loss to the GRPO objective with a fixed weight. \textbf{Skill-SD}~\cite{wang2026skill} conditions only the teacher on retrieved skills and transfers the resulting behavior to the student through an importance-weighted distillation loss. \textbf{RLSD}~\cite{yang2026rlsd} lets the outcome advantage determine the direction of each token update and uses the teacher--student log-probability gap only to rescale its magnitude, with a contribution that decays during training. \textbf{SDAR}~\cite{lu2026self} augments GRPO with an auxiliary distillation term, where a sigmoid gate on the detached teacher--student gap emphasizes the tokens endorsed by a skill-conditioned teacher. \textbf{StepOPSD}~\cite{zhang2026stepopsd} operates at the level of action steps, where a teacher that observes a successful trajectory from the same group reweights the step-level advantages according to its log-probability gap with the student. \textbf{OPID}~\cite{yang2026opid} uses an external LLM analyzer to summarize each on-policy trajectory into an episode-level skill and skills for its critical steps, routes the latter to the corresponding steps, and turns the resulting log-probability shift into a token-level advantage. \textbf{SEED}~\cite{wu2026seed} first fine-tunes the policy on skills that an external LLM writes for its trajectories, and then lets the policy itself summarize its rollouts into skills during RL training, distilling the skill-conditioned teacher.

\section{Hyperparameters}
\label{app:hyperparameters}

Table~\ref{tab:hyperparameters} lists the hyperparameters used in our experiments. The RSIPO-specific hyperparameters are shared across benchmarks and base models, except for the maximum length of the analysis output and the treatment of invalid skills: on search-based tasks, invalid skills are not penalized, and valid outputs instead receive a small format reward.

\begin{table*}[!t]
\centering
\setlength{\tabcolsep}{3pt}
\caption{\textbf{Results on ALFWorld and search-based agentic tasks with Qwen2.5-3B-Instruct.} We report success rates (\%) on ALFWorld and exact-match accuracy (\%) on the seven search datasets. Methods marked with * use skills at test time. The \sethlcolor{topcolor}\hl{\textbf{best}} and \sethlcolor{secondcolor}\hl{\mbox{\underline{second-best}}} results are highlighted.}
\label{tab:main_results_qwen25_3b}
\resizebox{\textwidth}{!}{%
\begin{tabular}{l*{15}{C}}
\toprule
\multirow{2}{*}[-0.6ex]{\textbf{Method}} & \multicolumn{7}{c}{\textbf{ALFWorld}} & \multicolumn{8}{c}{\textbf{Search-Based Tasks}} \\
\cmidrule(lr){2-8} \cmidrule(lr){9-16}
& \textbf{Pick} & \textbf{Look} & \textbf{Clean} & \textbf{Heat} & \textbf{Cool} & \textbf{Pick2} & \textbf{Avg} & \textbf{NQ} & \textbf{Triv} & \textbf{Pop} & \textbf{Hotp} & \textbf{2Wk} & \textbf{MuS} & \textbf{Bam} & \textbf{Avg} \\
\midrule
\rowcolor{gray!10} \multicolumn{16}{c}{\textit{Qwen2.5-3B-Instruct}} \\
Vanilla
  & 44.4 & 11.1 & 6.2 & 15.4 & 28.6 & 12.5 & 21.9
  & 24.6 & 48.1 & 31.0 & 26.3 & 25.3 & 7.2 & 59.7 & 31.7 \\
Skill-Prompt*
  & 51.7 & 66.7 & 48.4 & 0.0 & 4.3 & 10.0 & 28.9
  & 23.7 & 46.2 & 30.6 & 24.4 & 22.1 & 7.5 & 12.5 & 23.9 \\
OPSD
  & 48.8 & 41.7 & 16.7 & 0.0 & 15.8 & 16.7 & 28.1
  & 0.1 & 0.1 & 0.1 & 0.0 & 0.0 & 0.0 & 0.0 & 0.0 \\
GRPO
  & 91.2 & 62.5 & \cellcolor{secondcolor}\underline{96.2} & 61.9 & 65.0 & 47.4 & 75.0
  & 39.3 & 60.6 & 41.1 & 37.4 & 34.6 & 15.4 & 26.4 & 36.4 \\
Skill-GRPO
  & 88.9 & 71.4 & 58.8 & 70.6 & 40.7 & 29.2 & 60.2
  & 43.5 & 58.8 & 43.0 & 36.8 & 32.2 & 11.7 & 12.5 & 34.1 \\
Skill-GRPO*
  & 94.3 & 57.1 & \cellcolor{topcolor}\textbf{100.0} & 66.7 & 73.1 & 57.1 & 80.5
  & 44.3 & 59.6 & 44.3 & 39.0 & 36.1 & 14.5 & 14.9 & 36.1 \\
GRPO+OPSD
  & \cellcolor{topcolor}\textbf{100.0} & \cellcolor{secondcolor}\underline{82.4} & 85.7 & 75.0 & 70.0 & 60.0 & 81.2
  & \cellcolor{secondcolor}\underline{44.9} & 61.2 & 45.2 & 40.4 & 38.5 & 16.0 & 66.1 & 44.6 \\
Skill-SD
  & 88.2 & 50.0 & \cellcolor{secondcolor}\underline{96.2} & 52.4 & 65.0 & 57.9 & 73.4
  & 44.4 & 60.4 & 44.0 & 39.5 & 40.4 & 15.4 & 64.9 & 44.1 \\
RLSD
  & 87.9 & 75.0 & 90.9 & 75.0 & 73.1 & 68.4 & 79.7
  & 41.5 & 58.6 & 42.3 & 40.4 & 40.2 & \cellcolor{secondcolor}\underline{16.8} & 66.9 & 43.8 \\
SDAR
  & \cellcolor{secondcolor}\underline{97.1} & 62.5 & \cellcolor{topcolor}\textbf{100.0} & 61.9 & 75.0 & \cellcolor{secondcolor}\underline{84.2} & 84.4
  & 44.8 & 58.1 & 44.3 & 38.6 & 36.2 & 15.7 & 66.1 & 43.4 \\
StepOPSD
  & 82.4 & 66.7 & 82.6 & 52.2 & 73.7 & 75.0 & 73.4
  & 43.6 & 61.2 & 43.8 & 39.2 & 38.1 & 15.8 & 64.5 & 43.7 \\
OPID
  & 92.7 & \cellcolor{topcolor}\textbf{100.0} & 88.9 & 70.0 & \cellcolor{secondcolor}\underline{84.2} & 70.0 & 84.3
  & \cellcolor{topcolor}\textbf{45.9} & \cellcolor{secondcolor}\underline{61.4} & 45.7 & \cellcolor{secondcolor}\underline{40.7} & 38.8 & 16.4 & 66.1 & 45.0 \\
SEED
  & \cellcolor{topcolor}\textbf{100.0} & \cellcolor{topcolor}\textbf{100.0} & \cellcolor{topcolor}\textbf{100.0} & \cellcolor{topcolor}\textbf{100.0} & 70.6 & 80.0 & \cellcolor{secondcolor}\underline{91.8}
  & 44.3 & \cellcolor{topcolor}\textbf{61.9} & \cellcolor{secondcolor}\underline{47.2} & \cellcolor{topcolor}\textbf{41.0} & \cellcolor{secondcolor}\underline{42.0} & \cellcolor{topcolor}\textbf{16.9} & \cellcolor{topcolor}\textbf{67.7} & \cellcolor{topcolor}\textbf{45.7} \\
\textbf{RSIPO}
  & \cellcolor{secondcolor}\underline{97.1} & \cellcolor{topcolor}\textbf{100.0} & \cellcolor{topcolor}\textbf{100.0} & \cellcolor{secondcolor}\underline{93.8} & \cellcolor{topcolor}\textbf{92.0} & \cellcolor{topcolor}\textbf{91.7} & \cellcolor{topcolor}\textbf{95.7}
  & 39.0 & 61.1 & \cellcolor{topcolor}\textbf{49.8} & 39.6 & \cellcolor{topcolor}\textbf{43.2} & \cellcolor{topcolor}\textbf{16.9} & \cellcolor{secondcolor}\underline{67.3} & \cellcolor{secondcolor}\underline{45.3} \\
\bottomrule
\end{tabular}}
\end{table*}

\section{Training Algorithm of RSIPO}
\label{app:algorithm}

Algorithm~\ref{alg:rsipo} summarizes one training step of RSIPO. The policy first samples $G$ rollouts for each task and writes a skill for each of them, which yields the RLVR loss in Eq.~\eqref{eq:rlvr} and the OPSD terms in Eq.~\eqref{eq:opsd_term}. RSIPO then runs the execution check for every task. It splits the rollouts into an evidence set and a control set, rewrites a skill from an evidence trajectory, and replays the skill-enhanced teacher $K$ times from the same initial state to measure the replay gain $\Delta_x$ in Eq.~\eqref{eq:gain}. The gains of all tasks update the type-level statistics in Eq.~\eqref{eq:ema}, from which RSIPO computes the shrunk gains $\hat g_x$, the skill rewards $r_x$, and the trust coefficients $\omega_x$. Finally, the policy is updated with the overall objective in Eq.~\eqref{eq:total}, so that skill generation and task execution are optimized in the same step.

\begin{algorithm}[!ht]
\small
\caption{One training step of RSIPO}
\label{alg:rsipo}
\begin{algorithmic}[1]
\REQUIRE policy $\pi_\theta$, task batch $X$, type statistics $\{\mathbf{m}_c\}$
\STATE Sample $G$ rollouts per task; the policy writes a skill for each rollout
\STATE Compute $\mathcal{L}_{\text{RLVR}}$ (Eq.~\eqref{eq:rlvr}) and the OPSD terms $\ell_{t,j}$ (Eq.~\eqref{eq:opsd_term})
\FOR{each task $x\in X$}
  \STATE Randomly split its rollouts into $\mathcal{E}_x$ and $\mathcal{C}_x$
  \STATE Rewrite $s_x$ from a trajectory in $\mathcal{E}_x$, preferring a successful one
  \IF{$s_x$ is valid}
    \STATE Replay $\pi_\theta(\cdot\mid\cdot\oplus s_x)$ $K$ times from the initial state of $x$
    \STATE Compute $\Delta_x$ (Eq.~\eqref{eq:gain}) and its noise variance $v_x$
  \ENDIF
\ENDFOR
\STATE Update $\{\mathbf{m}_c\}$ (Eq.~\eqref{eq:ema}) and compute $\hat g_x$ (Eq.~\eqref{eq:shrink})
\STATE Compute $r_x$ (Eq.~\eqref{eq:reward}) and $\omega_x$ (Eq.~\eqref{eq:trust})
\STATE Update $\pi_\theta$ with $\mathcal{L}$ (Eq.~\eqref{eq:total})
\end{algorithmic}
\end{algorithm}

\definecolor{topcolor}{RGB}{252,236,196}
\definecolor{secondcolor}{RGB}{223,235,253}
\newcolumntype{C}{>{\centering\arraybackslash}p{0.058\textwidth}}

\begin{figure*}[t]
\centering
\includegraphics[width=\textwidth]{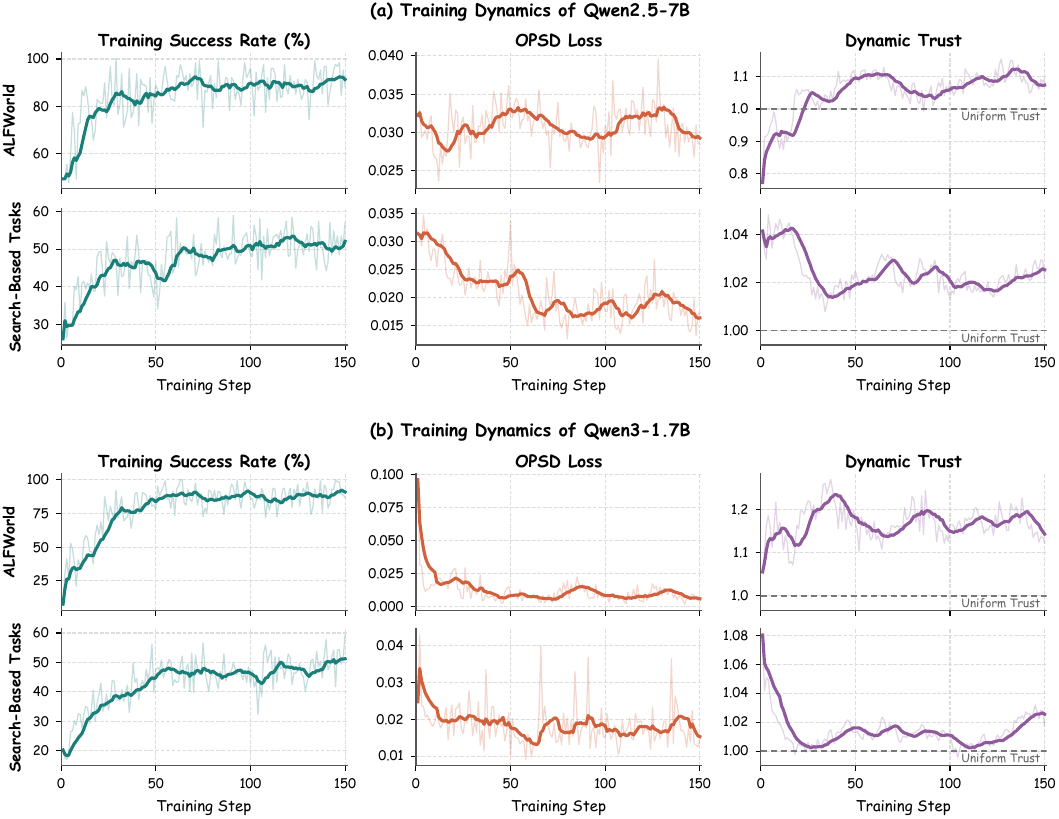}
\caption{Training dynamics of RSIPO with \textbf{(a)} Qwen2.5-7B and \textbf{(b)} Qwen3-1.7B on both benchmarks. Thin lines show per-step values and thick lines show 10-step moving averages.}
\label{fig:dynamics_appendix}
\end{figure*}

\section{Additional Experiments}

\subsection{Main Results on Qwen2.5-3B-Instruct}

Table~\ref{tab:main_results_qwen25_3b} compares RSIPO with all baselines using Qwen2.5-3B-Instruct. On ALFWorld, RSIPO achieves the best average success rate of 95.7\%, outperforming the strongest baseline, SEED, by 3.9\% and GRPO by 20.7\%. On search-based tasks, it ranks second with an average accuracy of 45.3\%, only 0.4\% below SEED.

\subsection{Training Dynamics of Other Base Models}
\label{app:dynamics}

Figure~\ref{fig:dynamics_appendix} shows the training dynamics of RSIPO with Qwen2.5-7B and Qwen3-1.7B, which follow trends similar to those of Qwen2.5-3B in Figure~\ref{fig:dynamics_ablation} (a). The training success rate rises steadily on both benchmarks, the OPSD loss decreases on search-based tasks, and the trust stays above one on ALFWorld for most of training and close to one on search-based tasks. For Qwen2.5-7B, the trust on ALFWorld starts below one because the teacher initially underperforms the student, and exceeds one after about 20 steps. For Qwen3-1.7B, the OPSD loss on ALFWorld drops to around 0.01.

\section{Prompts}
\label{app:prompts}

\begin{figure*}[!t]
\centering
\includegraphics[width=\textwidth]{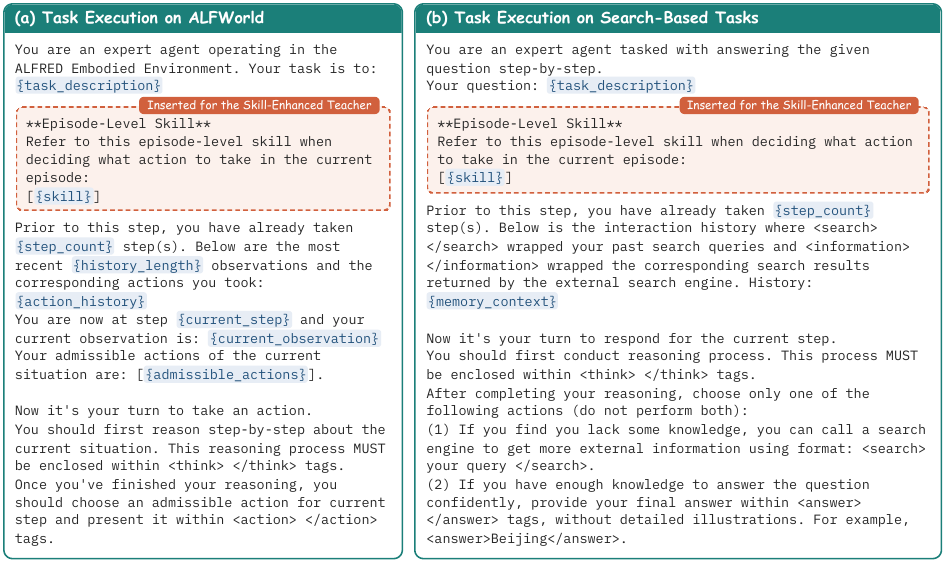}
\caption{Task-execution prompts on \textbf{(a)} ALFWorld and \textbf{(b)} search-based tasks. The dashed block is inserted only for the skill-enhanced teacher, and the history is omitted at the first turn.}
\label{fig:prompt_execution}
\end{figure*}

\begin{figure*}[!t]
\centering
\includegraphics[width=\textwidth]{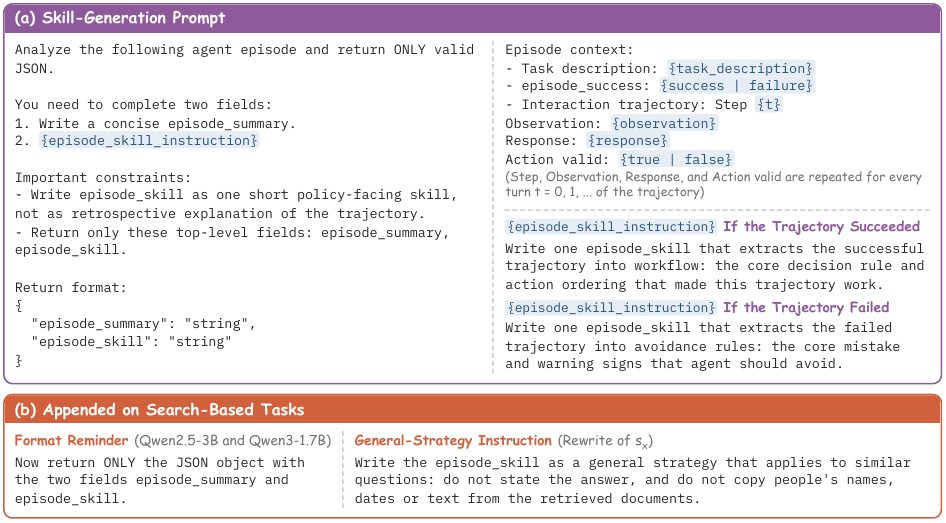}
\caption{Skill-generation prompt. \textbf{(a)} The analysis template, whose instruction for the episode skill depends on the outcome of the trajectory. \textbf{(b)} Additions on search-based tasks.}
\label{fig:prompt_skill}
\end{figure*}

\begin{figure*}[!t]
\centering
\includegraphics[width=\textwidth]{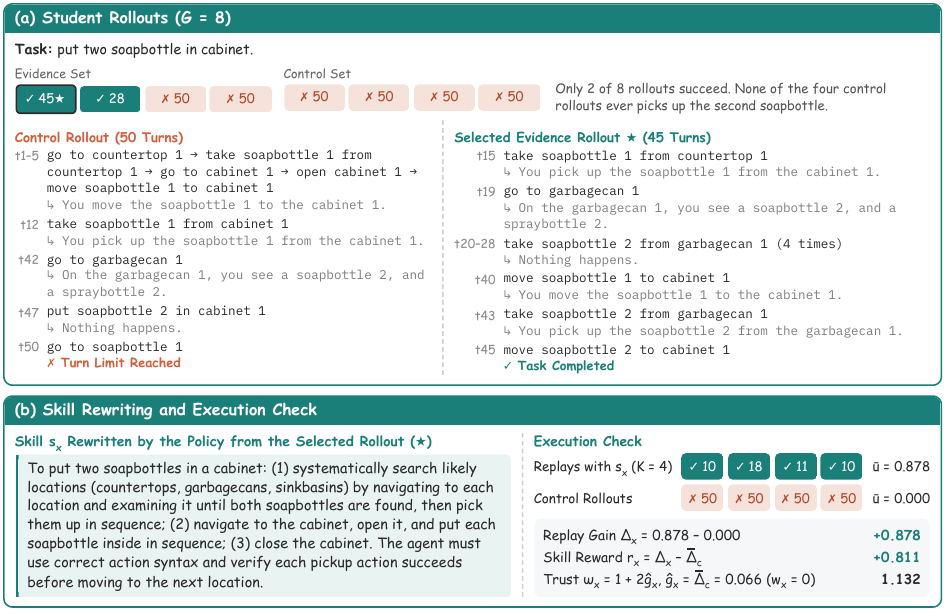}
\caption{A helpful skill written by Qwen2.5-7B for a pick-two task on ALFWorld.}
\label{fig:case_alfworld}
\end{figure*}

\begin{figure*}[!t]
\centering
\includegraphics[width=\textwidth]{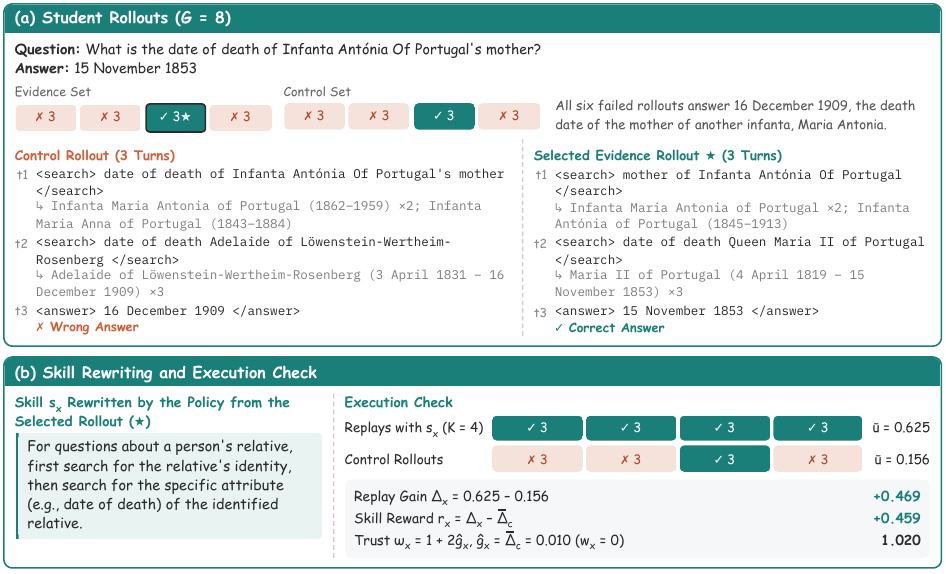}
\caption{A helpful skill written by Qwen2.5-7B for a 2WikiMultihopQA question on search-based tasks.}
\label{fig:case_search}
\end{figure*}

Figure~\ref{fig:prompt_execution} shows the prompts for task execution. At each turn, the student receives the task, the most recent turns of interaction, and the current observation, and the history is omitted at the first turn. The skill-enhanced teacher, which scores the student tokens in OPSD and performs the replays in the execution check, receives the same context with the episode-level skill inserted after the task description. Figure~\ref{fig:prompt_skill} shows the analysis template (Following SEED~\cite{wu2026seed}), with which the policy writes the skills for OPSD and rewrites $s_x$ from the selected evidence trajectory. The instruction for the episode skill depends on whether the trajectory succeeded. On search-based tasks, the rewrite prompt additionally asks for a general strategy that neither states the answer nor copies the retrieved documents, so that a high replay gain cannot be obtained by leaking the answer to the teacher, and for Qwen2.5-3B and Qwen3-1.7B the output format is restated after the trajectory.

\section{Case Study}
\label{app:case_study}

Figures~\ref{fig:case_alfworld} and~\ref{fig:case_search} show the execution check on a helpful skill for each benchmark, both taken from late training steps of Qwen2.5-7B. In both figures, \ding{51} and \ding{55} mark successful and failed episodes together with their numbers of turns, and \ding{72} marks the selected evidence rollout. The trajectory excerpts list the actions at selected turns $t$, each followed by the next observation on ALFWorld or by the titles of the retrieved documents on search-based tasks.

\paragraph{ALFWorld.} In Figure~\ref{fig:case_alfworld}, the student often fails to put two soapbottles into a cabinet: only two of the eight rollouts succeed, and none of the four control rollouts ever picks up the second soapbottle. As the excerpts show, the student takes the first soapbottle out of the cabinet again after placing it. In the selected evidence rollout, it finds the second soapbottle in the garbage can at turn 19 while still holding the first, and its attempts to pick up the second have no effect until it puts the first back at turn 40. From this 45-turn success, the policy writes a skill that lists garbage cans among the places to search and asks the agent to verify that each pickup succeeds. With this skill, all four replays succeed within 18 turns while all control rollouts fail, giving $\Delta_x=0.878$ and $r_x=0.811$.

\paragraph{Search-Based Tasks.} In Figure~\ref{fig:case_search}, the question asks for the death date of the mother of Infanta Ant\'onia of Portugal. The six failed rollouts search for the death date directly, retrieve another infanta, Maria Antonia, and answer the death date of her mother. The selected evidence rollout instead searches for the mother first, identifies Maria II, and then searches for her death date. The policy turns this rollout into a general strategy that contains neither a name nor the answer: identify the relative first, then search for the requested attribute. With this skill, all four replays answer correctly, whereas only one of the four control rollouts does, giving $\Delta_x=0.469$ and $r_x=0.459$.

\end{document}